\documentclass[11pt]{article}

\usepackage{caption}

\usepackage{changes}

\usepackage[a4paper]{geometry}
\usepackage{latexsym}
\usepackage{authblk}
\usepackage{clic2017}
\usepackage{times}
\usepackage{url}
\usepackage{latexsym}

\usepackage{paralist}
\usepackage{amsmath}
\usepackage{array}
\usepackage{tabularx}
\usepackage{multirow}
\usepackage{graphicx}
\usepackage{amsfonts}
\usepackage{amssymb}
\usepackage{url}
\usepackage{tcolorbox}
\usepackage{enumitem}
\usepackage{graphicx}
\usepackage{multirow, makecell}
\usepackage{lingmacros}
\usepackage{bbm}
\usepackage{booktabs}
\usepackage[T1]{fontenc}
\usepackage[outline]{contour}
\usepackage{xcolor}
\contourlength{0.2pt}
\contournumber{55}%
\newcolumntype{C}{>{\centering\arraybackslash}X}

\usepackage[caption=false]{subfig}

\newcolumntype{S}{>{\arraybackslash}m{6cm}}
\newcolumntype{F}{>{\arraybackslash}m{2cm}}
\newcolumntype{P}{>{\arraybackslash}m{3cm}}

\usepackage[T1]{fontenc}
\usepackage[outline]{contour}
\usepackage{xcolor}
\contourlength{0.2pt}
\contournumber{55}%

\usepackage[utf8]{inputenc}
\usepackage{amsmath}
\usepackage{fourier}
\usepackage{esvect}
\usepackage{scalerel}

\usepackage{times}
\usepackage{helvet}
\usepackage{courier}
\usepackage{graphicx} 
\usepackage{fancyhdr}
\usepackage{etoolbox}
\usepackage[export]{adjustbox}
\usepackage{float}
\usepackage{amsmath}
\usepackage{mathrsfs}
\title{Insights into Schizophrenia: Leveraging Machine Learning for Early Identification via EEG, ERP, and Demographic Attributes}

\author[1,2,3]{\textbf{Sara Alkhalifa}}
\affil[1]{Imam Abdulrahman Bin Faisal University, Kingdom of Saudi Arabia}
\affil[2]{Alhassan School, Kingdom of Saudi Arabia}
\affil[3]{Summer Research Program (H Xplore) 2021, King Fahd University of Petroleum and Minerals}

\date{}

\begin{document}
\maketitle
\begin{abstract}

The research presents a machine learning (ML) classifier designed to differentiate between schizophrenia patients and healthy controls by utilising features extracted from electroencephalogram (EEG) data, specifically focusing on event-related potentials (ERPs) and certain demographic variables. The dataset comprises data from 81 participants, encompassing 32 healthy controls and 49 schizophrenia patients, all sourced from an online dataset. After preprocessing the dataset, our ML model achieved an accuracy of 99.930\%. This performance outperforms earlier research, including those that used deep learning methods. Additionally, an analysis was conducted to assess individual features' contribution to improving classification accuracy. This involved systematically excluding specific features from the original dataset one at a time, and another technique involved an iterative process of removing features based on their entropy scores incrementally. The impact of these removals on model performance was evaluated to identify the most informative features.

\end{abstract}

\section{Introduction}

It has been demonstrated that various animals, including humans, can suppress or minimise the sensory consequences of their actions \cite{adolphs2013biology}. This is achieved through the corollary discharge (CD) mechanism, wherein a duplicate of the motor command is sent to the muscles for movement. However, this analogy, or copy, does not initiate movement; instead, it alerts other brain parts to the impending action \cite{mathalon2008corollary}. For instance, when you speak, your auditory cortex exhibits a reduced response to the expected sound of your voice. People living with schizophrenia often struggle to predict sensations and events, leading to misinterpretations. Impairment in their predictive mechanisms may cause anticipated sensations to be given undue prominence \cite{Kapur2003}. This may be the reason why SZ patients often get misdiagnosed. 

Schizophrenia, a complex mental disorder with an unknown aetiology, affects approximately 1\% of the global population and can lead to cognitive impairments, hallucinations, delusions, and debilitating brain and behaviour disorders impacting thought processes, emotions, and actions \cite{psychiatryschizophrenia}. Symptoms often emerge early in life and may deteriorate over time, spanning decades \cite{brainfoundation_2019}. One possible explanation for specific schizophrenia symptoms is that disruptions in the corollary discharge process within the nervous system make it challenging for patients to distinguish between internally and externally generated stimuli \cite{mathalon2008corollary}. Therefore, investigating this process and its correlation with the illness's symptoms could enhance our understanding of abnormal brain processes in individuals with this diagnosis \cite{delisi2022understanding}.

In recent times, scientists have employed machine learning (ML) and artificial intelligence to detect, monitor, and predict various diseases, including schizophrenia, owing to the high performance of these techniques in uncovering associations between symptoms and diseases \cite{shatte2019machine}.

An early diagnosis of schizophrenia facilitates timely and tailored therapy for symptom relief. However, this poses a challenge for psychiatrists who often need to make decisions on a case-by-case basis \cite{haefner2006early}. Misdiagnosing early symptoms or experiencing failure in the initially prescribed medication or treatment can result in the loss of the optimal window for disease control and treatment, potentially leading to adverse side effects for the patient \cite{kunzmann_2021}. The significant challenge lies in the difficulty of differentiating between rational and irrational behaviour \cite{rosenhan1973being,coulter2019specialized}—an issue that machine learning can potentially address by serving as a statistical tool to aid psychiatrists in identifying schizophrenia and determining the most suitable treatment for each patient \cite{sun2021hybrid}.
   
 \subsection{Research aims}
\label{ssec:research-aims}

The focus of this study is to reduce the misdiagnosis of schizophrenia patients by classifying EEG signals using three machine learning models.

\begin{itemize}
    \item The primary objective is to develop a rigorous and systematic research technique that may effectively enhance the accuracy of schizophrenia diagnosis by utilising machine learning models and comprehensive data analysis.
    
    \item This study aims to establish a systematic and rigorous research methodology to improve the precision of schizophrenia diagnosis. This will be achieved by employing machine learning models and doing thorough data analysis.

    \item  By examining the relationships between sociodemographic characteristics and neurophysiological markers, notably event-related potentials (ERPs) and electroencephalogram (EEG) signals, a thorough understanding of the symptoms of schizophrenia may be achieved.
    
    \item Assess the influence of different datasets on the diagnostic accuracy of machine learning models in relation to schizophrenia.

    \item The present research aims to analyse and identify patterns and trends in the performance of machine learning models that have been trained using varied datasets.
\end{itemize}
   
 \subsection{Research questions}
\label{ssec:research-questions}

  This research uses machine learning classifiers to distinguish between EEG signals from schizophrenia patients and healthy controls. The main goal of this study's machine learning challenge is to uncover valid EEG biomarkers for diagnosing schizophrenia from a set of features. The ultimate goal is to turn the research findings into a valuable tool for physicians in therapeutic situations.
This research is answering the following research questions:

\begin{itemize}
    
    \item To what extent can research approaches contribute to the precision of schizophrenia diagnosis, and how might their use increase the accuracy of the diagnostic process?
    
    \item To what degree can demographic variables, including age, gender, and socio-economic position, affect the precision of machine learning models in forecasting the likelihood of schizophrenia risk using EEG and ERP data?
    
    \item Can the augmentation of input data impact machine learning outcomes, either positively or negatively?

    \item How may the discoveries of this investigation enhance the progress of targeted interventions and personalised treatment approaches for individuals diagnosed with schizophrenia via the utilisation of machine learning analyses?

    \item What are the most essential features that can affect the ML's accuracy?

\end{itemize}

\subsection{Contributions}
This thesis highlights the subsequent contributions:
\begin{itemize}
    \item This study introduces a machine learning methodology designed for the analysis of electroencephalography (EEG), event-related potential (ERP) data, and demographic factors to improve the precision of schizophrenia detection.

    \item  This study offers mental health professionals and physicians a comprehensive diagnostic framework that facilitates informed treatment decisions for schizophrenia while reducing the likelihood of misdiagnosing other conditions.

    \item This research work significantly adds to the area by investigating the effects of machine learning techniques to analyse a dataset, including neurophysiological signals and demographic data. Improving the precision of schizophrenia diagnosis has the potential to enhance the efficacy of diagnoses with schizophrenia. 
    
    \item This study improves clinical diagnosis accuracy by utilising ML for the early detection of schizophrenia. Attributed to the ML model's exceptional accuracy, it may cause a substantial reduction in patients receiving erroneous diagnoses.
    
    \item The present study investigates the adaptability and robustness of machine learning models across various demographic contexts. The principal aim of this endeavour is to provide an academic contribution to the development of diagnostic instruments for schizophrenia that are universally applicable.

    \item The objective of this paper is to identify the most critical characteristics of EEG signals in order to diagnose schizophrenia.

\end{itemize}
\label{ssec:contributions}

\subsection{Paper structure}

This paper is organized as follows: Section 2.1 presents an overview of research pertaining to EEG. Section 2.2 examines machine learning models in psychiatry, reviewing pertinent literature, highlighting the importance of these models in mental health diagnosis, and setting the baseline for evaluation. The following sections concentrate on the methodology. Section 3.1 outlines the data collection process, Section 3.2 discusses the application of machine learning models for classification, and Section 3.3 addresses the attribute selection methodology. The sections delineate the researcher's methodology, encompassing data acquisition and the practical implementation of machine learning models.

\section{Related work}
\label{sec:rw}

\subsection{Overview of research in EEG}

\begin{table*}[h] 
\centering
\begin{adjustbox}{max width=0.98\textwidth}
\begin{tabular}{|l|S|F|S|F|}
\hline
\hline
        \textbf{Author(year)} & \textbf{EEG dataset} & \textbf{Method} & \textbf{Features and Validation method} & \textbf{Accuracy} \\ \hline \hline 
        \cite{johannesen2016machine} & 57 schizophrenia patients and 24 normal subjects & SVM & Absolute power analysis & 83.33\% \\ \hline
        \cite{johannesen2016machine} & 40 schizophrenia patients and 12 healthy controls  & SVM & Morlet continuous wavelet transform & 87\% \\ \hline
        \cite{jeong2017classifying} & 30 schizophrenia patients and 15 controls & SKLDA & Mean subsampling technique & Over 98\% \\ \hline

        \cite{piryatinska2017binary} & 45 boys suffering from schizophrenia and 39 healthy boys & RF & ?-complexity of a continuous vector function & 85.3\% \\ \hline
        \cite{chu2017analysis} & 10 normal and 17 markedly ill schizophrenic patients & SVM & ApEn & 81.5\% \\ \hline
        \cite{alimardani2018db} & 26 subjects with schizophrenia and 27 patients with BMD & NN & DB-FFR & 87.51\% \\ \hline
        \cite{alimardani2018classification} & 23 bipolar disorder and 23 schizophrenia subjects & KNN & SSVEP SNR & 91.3\% \\ \hline
      
        \cite{phang2019multi} & 45 schizophrenia patients and 39 healthy controls & DNN-DBN & Vector-autoregression-based directed connectivity (DC), graph-theoretical complex network (CN) & 95\% \\ \hline
        \cite{phang2019multi} & 45 schizophrenia patients and 39 healthy controls & MDC-CNN & Directed connectivity measures (VAR coefficients and PDCs) and topological CN measures & 91.69\% \\ \hline
        \cite{oh2019deep} & 14 healthy subjects and 14 SZ patients & CNN & -- & 98.07\% \\ \hline
        \cite{sun2021hybrid} & 54 patients with schizophrenia and 55 healthy controls & CNN+LSTM & FuzzyEn & 99.22\% \\ \hline
        \cite{borgwardt2013distinguishing} & 22 HC, 23 FEP & SVM (nonlinear) & Nested cross-validation & 86.7\% \\ \hline
        \cite{davatzikos2005whole} & 79 HC, 69 SZ & SVM (nonlinear) & LOOCV & 81.1\% \\ \hline
        \cite{fan2006compare} & Female sample: 38 HC, 23 SZ & SVM (nonlinear) & LOOCV & 91.8\% \\ \hline
        ~ & Male sample: 41 HC, 46 SZ & ~ & ~ & 90.8\% \\ \hline
        \cite{nieuwenhuis2012classification} & Training sample: 111 HC, 128 SZ & SVM (linear) & LOOCV & 71.4\% \\ \hline
        ~ & Validation sample: 122 HC, 155 SZ & ~ & Independent sample & 70.4\% \\ \hline
        \cite{dubb2004regional} & 46 HC, 46 SZ & SVM & LOOCV & 70.7\% \\ \hline
        \cite{zanetti2013neuroanatomical} & 62 HC, 62 FE & SVM (nonlinear) & LOOCV & 73.4\% \\ \hline
        \cite{sun2009elucidating} & 36 HC, 36 ROS & Sparse multinomial & LOOCV & 86.1\% \\ \hline
        ~ & ~ & logistic regression & ~ & ~ \\ \hline
        \cite{greenstein2012using} & 99 HC, 98 SZ & Random forests & Out-of-bag (33\% left out at each tree) & 73.6\% \\ \hline
        \cite{karageorgiou2011neuropsychological} & 47 HC, 28 ROS & LDA & LOOCV & 64.3\% SEN \\ \hline
        ~ & ~ & ~ & ~ & 76.6\% SPC \\ \hline
        \cite{kasparek2011maximum} & 39 HC, 39 FEP & LDA & Jackknife (LOOCV) & 71.8\% \\ \hline
        \cite{kawasaki2007multivariate} & Training sample: 30 HC, 30 SZ & LDA & LOOCV & 76.7\% \\ \hline
        ~ & Held-out group: 16 HC, 16 SZ & ~ & Held-out group & 84.4\% \\ \hline
        \cite{nakamura2004multiple} & Female sample: 22 HC, 27 SZ & LDA & N/A & 81.6\% \\ \hline
        ~ & Male sample: 25 HC, 30 SZ & ~ & ~ & 80\% \\ \hline
\hline
\end{tabular}
\end{adjustbox}
\caption{Summary of Studies using ML methods to detect SZ patients. SEN: sensitivity, SPC: specificity}
\label{tab:related-table}
\end{table*}

Mental disorders like schizophrenia may continue to grow once it is started. Development might differ significantly for specific individuals. In early identification of the mental beginning and progressive stages, fast and effective therapy to prevent or reduce future illness degeneration is crucial. The Mental Disorders Diagnostic and Statistical Manual (DSM-5) offers diagnostic information on various mental illnesses. The traditional image technique includes asking the patient about the symptoms and life of the disease in multiple ways.

Clinical measures like the Positive and Negative Syndrome Scale (PANSS) may evaluate the strength of the symptoms caused by misdiagnosis. Thus, training EEG and other data machine learning models is a helpful way to reduce schizophrenia misdiagnosis, which can significantly improve the clinical condition.
  
It was reported that EEG data from a single electrode had been claimed to be adequate to diagnose schizophrenia with a modification in time frequency \cite{dvey2015schizophrenia}. It has also been shown that EEG theta frequency activity in schizophrenic individuals has reduced functional connection \cite{konig2001decreased}.

\subsection{Machine Learning Model in Psychiatry}

According to new research, integrating machine learning to both source and sensor-level EEG information can enhance the accuracy of diagnosing schizophrenia (the features at the source level being spatial features related to the active brain region). EEG biomarkers, such as the amplitude and latency of event-related potentials (ERP), and EEG Trial Data with machine learning for classification between schizophrenia patients and healthy controls have also been used at the sensor level \cite{zhang2019eeg}. 

\section{Methodology}

   \begin{figure}[t]
    \footnotesize
 \begin{center} 
  \includegraphics[scale=0.40]{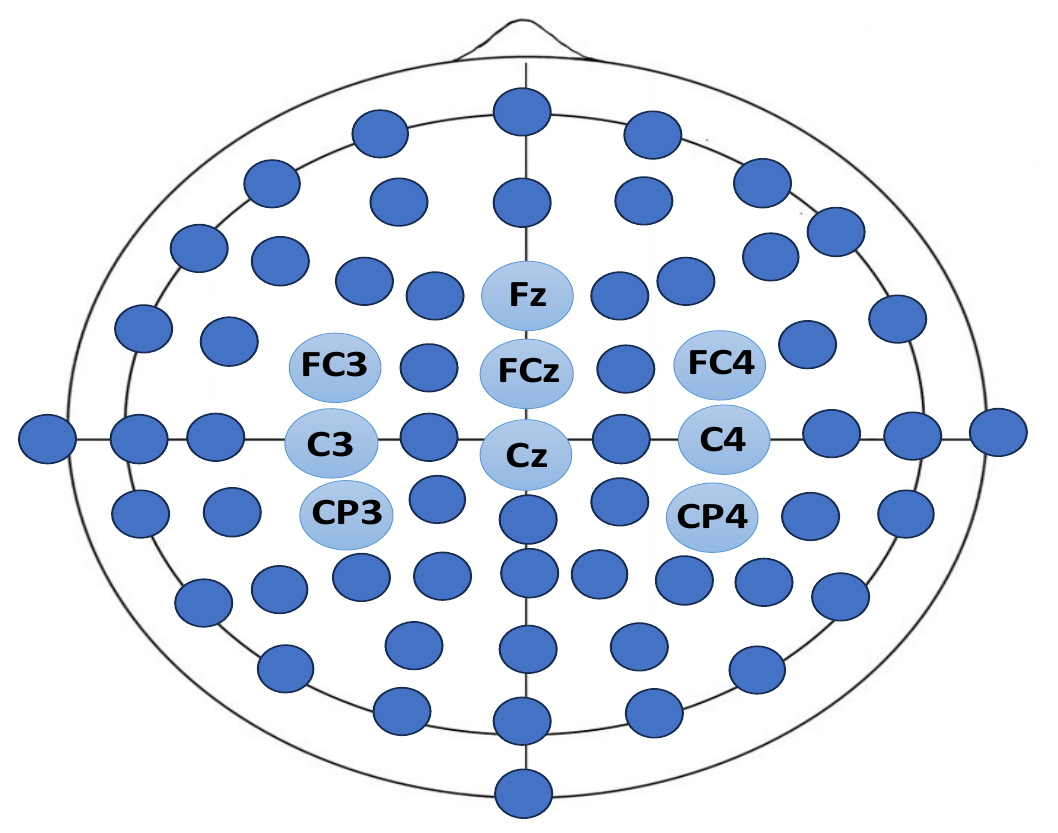} 
  \caption{An overview of EEG.} 
  \label{fig:model}
 \end{center}
\end{figure}
 
This section explains the online data collection process and demonstrates the diagnostic potential of EEG signals in identifying SZ. Next, the researcher explains the machine learning models and the measurements utilised for attribute selection in this study.  

\subsection{Data Collection}

  Data were collected from an online EEG data set and available publicly \cite{Ford2014}. Figure \ref{fig:model} contains the average event-related potentials (ERPs) for the nine electrode sites tested [Fz, FCz, Cz, FC3, FC4, C3, C4, CP3, CP4].
  
  
   Before a motor action, there is an efference copy of the motor order. This causes the expected sensation to be transmitted to the sensory cortex. These mechanisms serve animals to predict sensations, suppress responses to self-generated sensations, and process sensations efficiently and economically. People with schizophrenia have less talking activity and less speech sound suppression when they are interacting with others, which is consistent with issues with efference copy and corollary discharge.\cite{Ford2014}

   To investigate corollary discharge in individuals with schizophrenia and comparison controls, the data collector employed the following method: (1) pressing a button to generate a tone immediately, (2) passively listening to the tone, or (3) holding down the button without producing a tone to facilitate comparison between schizophrenia patients and controls. Researchers observed that controls exhibited suppression of the N100, a negative deflection in the EEG brain wave occurring 100 milliseconds post-sound onset, upon pressing a button to generate a tone; however, this suppression was absent in schizophrenia patients. This indicates a disruption in the efference copy mechanism, which typically aids the brain in predicting and mitigating the sensory outcomes of self-initiated actions. The study underlined the possible part faulty predictive coding could have in the sensory anomalies and symptoms experienced by schizophrenic people. The results confirm the theory that schizophrenia is connected to a basic deficiency in the brain's capacity to predict and interpret action outcomes, therefore contributing to the sensory and cognitive abnormalities unique of the condition. 
   
    The EEG data was obtained from 22 healthy controls and 36 individuals diagnosed with schizophrenia, in addition to 10 controls and 13 schizophrenia patients from a previous study \cite{Ford2014}. The study included a total of 81 participants, consisting of 32 people in good health and 49 people diagnosed with schizophrenia.

  
\subsection{CLASSIFICATION USING MACHINE LEARNING MODELS}

The researcher used a data set comprising EEG signals, ERP data, and demographic attributes, including gender, age, and education, to train a machine learning model. Three models, namely SVM, k-NN, and decision tree classification, were utilized to ascertain the highest accuracy. Additionally, the models were tested on both ERP-inclusive and ERP-exclusive data. The researcher also attempted to include it in machine learning models alone without amalgamating it with the remaining dataset. Furthermore, the researcher assessed the significance of the columns by employing two methods of column reduction. The first method involved incrementally removing one column at a time until only one column remained (incremental) according to the entropy. The second method involved removing one column at a time and reintroducing it to complete all columns' processes. 

  \begin{figure}[t]
    \footnotesize
 \begin{center} 
  \includegraphics[scale=0.4]{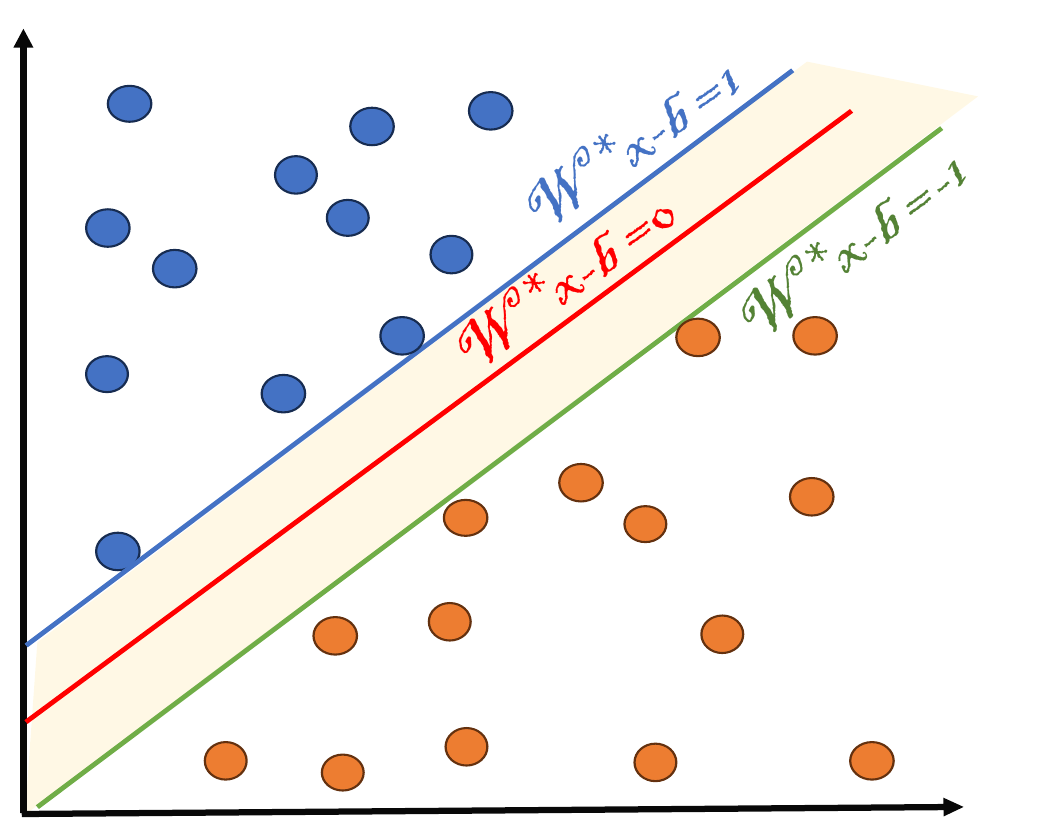} 
  \caption{ An overview of SVM.}
  \label{fig:SVM}
 \end{center}
\end{figure}

\begin{itemize}
    \item \textbf{SVMs} are linear models that handle classification and regression problems. This tool excels in dealing with both linear and nonlinear problems, as well as a variety of practical concerns. SVM is conceptually straightforward to build because its primary purpose is to segregate the supplied dataset as effectively as possible. SVM seeks the greatest marginal partition of the data into classes by generating lines. 
    SVM, on the other hand, may be expanded to handle nonlinear classification issues where the collection of data cannot be separated linearly, as demonstrated in \ref{fig:SVM}. Using kernel functions to map Samples in a high-dimensional feature space, resulting in linear classification. SVM algorithms use a set of mathematical functions defined as the kernel, so by using it, we can separate nonlinearly separable datasets which couldn't be solved by simple logistic regression. The most popular RBF kernel is the general kernel (based on the RBF kernel). Different SVM algorithms use other types of kernel functions. These functions can be linear, nonlinear, polynomial, radial basis function (RBF), and sigmoid. Introduce Kernel functions for sequence data, graphs, text, images, and vectors. The most used type of kernel function is RBF. Because it has localised and finite responses along the entire x-axis. The kernel functions return the inner product between two points in a suitable feature space. Thus, it defines a notion of similarity with a bit of computational cost, even in very high-dimensional spaces. RBF kernel can be considered a similarity measure, and distances can be regarded as a dissimilarity measure. RBF kernels are good approximations of K-nearest neighbours. Similar to K, sigma is used in K-NN. When K increases, overfitting occurs, and we try to minimise the loss function, which is identical to the sigma in the previous sentence. It's always preferable to use RBF kernel when uncertain of the best function for determining parameters. Equation is:
    
         \begin{equation}
                \begin{aligned}
                \mathcal{X}\left(\mathbb{X}_i, \mathbb{X}_j\right)=\exp \left(-\gamma\left\|\mathbb{X}_{i} -\mathbb{X}_j\right\|^2\right)
                \end{aligned}
    \end{equation}

    \begin{figure}[t]
        \footnotesize
         \begin{center} 
          \includegraphics[scale=0.4]{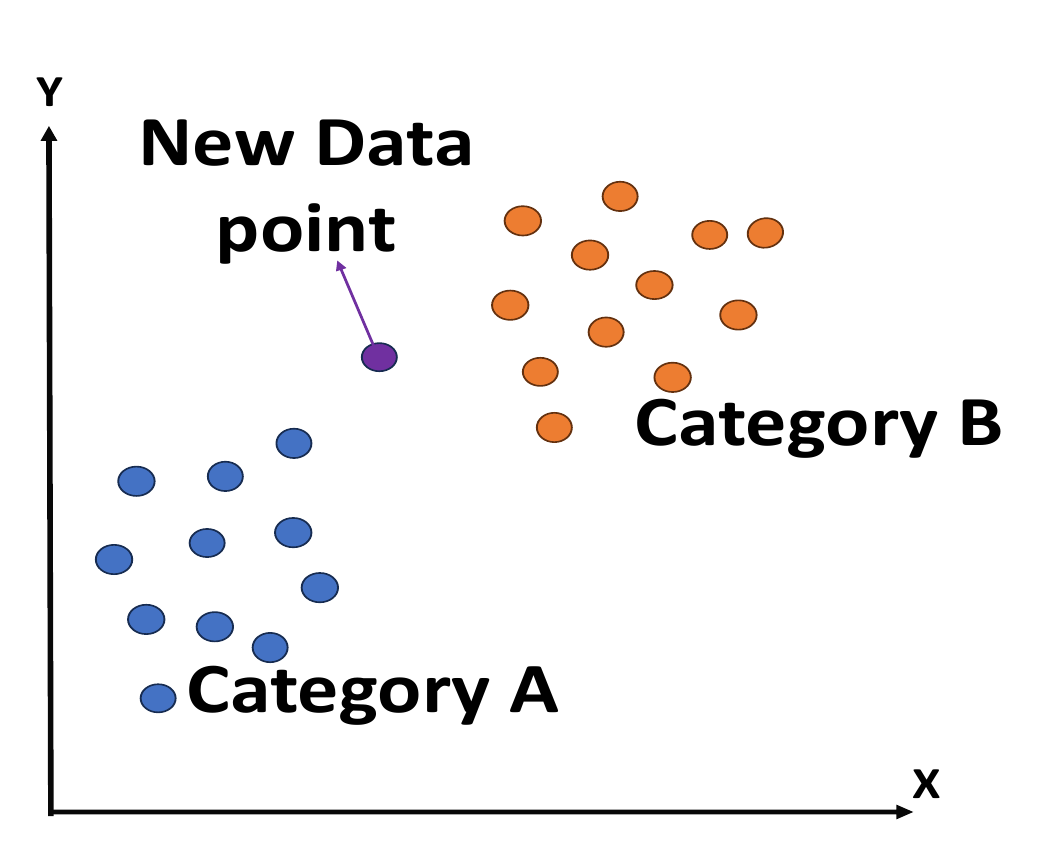}
          \caption{An overview of K-NN.}
          \label{fig:KNN}
         \end{center}
        \end{figure}
        
    \item \textbf{KNN} is often used as a non-parametric algorithm that is easy to learn, but it is also a relatively straightforward and easily adaptable learning process. Predicting the classification of a new sample point is done by looking at a database of previous sample points that have been classified. The data is used to construct a KNN model, which replaces the original data for classification. k is automatically determined, varies with different data types, and is optimal in classification accuracy. The construction of the model reduces the dependency on k and makes classification faster. Based on the following algorithm, we can explain how K-NN works:
    
    \begin{itemize}
        \item 	As a first step, choose a K for identifying neighbours.
        \item 	The next step is to compute Euclidean distances between neighbours K.
        \item 	The third step involves computing the Euclidean distance and determining the closest K neighbours.
        \item 	Next, count the data points for each category among these k neighbours.
        \item 	In the 5th Step, assign the data points to a category with the most neighbours.
        \item 	6th Step: The model has been completed.
    \end{itemize}
      Assume we have a new data point that needs to be assigned to a category. As illustrated in Figure \ref{fig:SVM}, it starts by deciding on the number of neighbours. The Euclidean distance between the data points will then be calculated. It can be calculated using the following equations:

         \begin{equation}
                \begin{aligned}
                \sqrt{({X_{2}-X_{1}})^2 +({Y_{2}-Y_{1}})^2}  
                \end{aligned}
    \end{equation}
      
      By calculating the Euclidean distance, we can get the nearest neighbours.

    \item DECISION TREE CLASSIFICATION
    
A decision tree-based architecture is employed in constructing regression and classification models. During this procedure, the dataset is progressively divided into ever smaller subgroups, leading to the gradual formation of a decision tree. The resultant structure is a comprehensive tree with decision nodes and leaf nodes. One example of a weather forecasting system that demonstrates division into distinct categories is Outlook, which classifies weather conditions into three branches: Sunny, Overcast, and Rainy. In classifications or judgments, leaf nodes serve as representations, exemplified by the category "Play." The root node is considered the most effective predictor inside a decision since it can handle categorical and numerical input. The user requests a discussion on the image labelled "Figure 4: Decision Tree."

        \begin{figure}[t]
        \footnotesize
         \begin{center} 
          \includegraphics[scale=0.4]{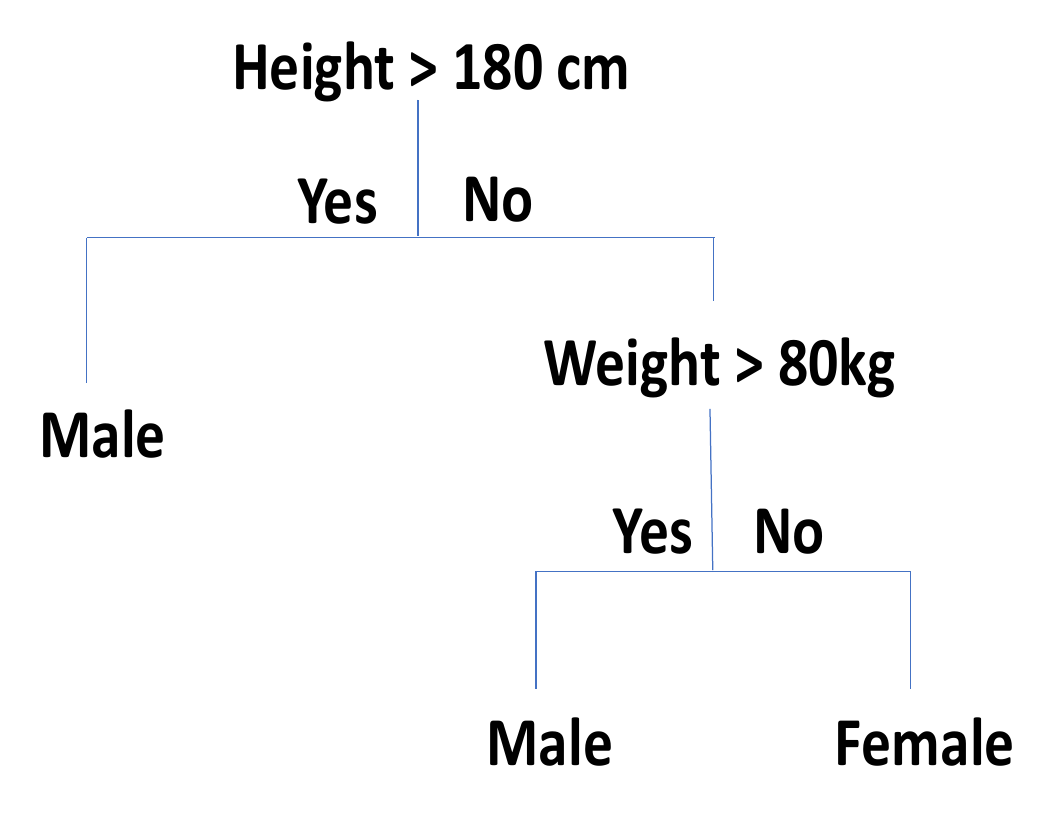}
          \caption{ An overview of DECISION TREE.}
          \label{fig:DECISION_TREE}
         \end{center}
        \end{figure}
    
    The following is a simple example in the form of a decision tree. A person is classified as male according to characteristics: taller than 180cm or shorter than 180cm but heavier than 80kg. Otherwise, she is classified as female. 
\end{itemize}

\subsection{Attribute Selection Measures}

When a dataset has N features, determining whether to structure them as internal nodes at the root or different levels of the tree might be difficult. The issue could not be addressed by picking a random node as the root because this would yield poor results and low precision. Therefore, researchers collaborated to tackle the attribute selection problem. Finally, they recommended that you use the following criteria:

\begin{itemize}
    \item \textbf{Entropy} measures how random the information being processed is. The higher the entropy, the more difficult it is to make inferences from the data. According to ID3, a branch with zero entropy is a leaf node, while a branch with an entropy greater than zero requires additional splitting. The researcher used this method to remove one column incrementally. The column with the highest entropy was deleted each time until only one remained. The entropy may be computed using the following formula:

    \begin{equation}
                \begin{aligned}
                E=-\sum_{i=1}^k p_i \log _2\left(p_i\right) 
                \end{aligned}
    \end{equation}
    
    Where S → Current state, and Pi → Probability of an event i of state S or Percentage of class i in a node of state S.
    
\end{itemize}



\section{Results}\label{sec:results}

\begin{table*}[h] 
\centering
\begin{adjustbox}{max width=0.96\textwidth}
\begin{tabular}{|l|l|l|l|}
    \hline 
        \textbf{Models} & \textbf{ERP} & \textbf{EEG \& demographic} & \textbf{ALL} \\ \hline \hline
        SVM & -- & 72.98\% & 80.78\% \\ \hline
        Decision Tree Classification & 79.14\% & 97.18\% & 99.93\% \\ \hline
        K-NN & 93.33\% & 74.32\% & 87.95\% \\ 
\hline
\end{tabular}
\end{adjustbox}
\caption{MACHINE LEARNING MODELS CLASSIFICATION ACCURACY RESULTS }
\label{tab:results-table}
\end{table*}

Table \ref{tab:results-table} shows the machine learning models used in this study (more information on the models can be found in the Methodology), as well as their accuracy for each dataset group: ERP (Event-Related Potential) and EEG (Electroencephalogram), merged with the Demographic (age, gender, education) dataset. Finally, all the datasets were combined in Excel and used in a machine learning model in Google Colap.  

As you can see in Table \ref{tab:results-table}, SVM (Sport Vector Machine) has the lowest accuracy compared to the other model’s accuracies, and the accuracy in all the data using the SVM model was the lowest, at 72.98\% using EEG and demographic  merged data. On the other hand, the highest accuracy was 99.93\% using all the datasets in Decision Tree Classification.

The researcher evaluated the features' importance by utilising two-feature reduction techniques to further this research analysis. Based on the entropy principle, the initial approach systematically eliminated one feature at a time until only one remained (incremental). The second approach entailed systematically eliminating one feature at a time and reintegrating it to ensure the completion of all feature procedures.

\begin{itemize}
\item\textbf{One feature-out experiment}


\begin{figure}[t]
        \footnotesize
         \begin{center} 
          \includegraphics[scale=0.25]{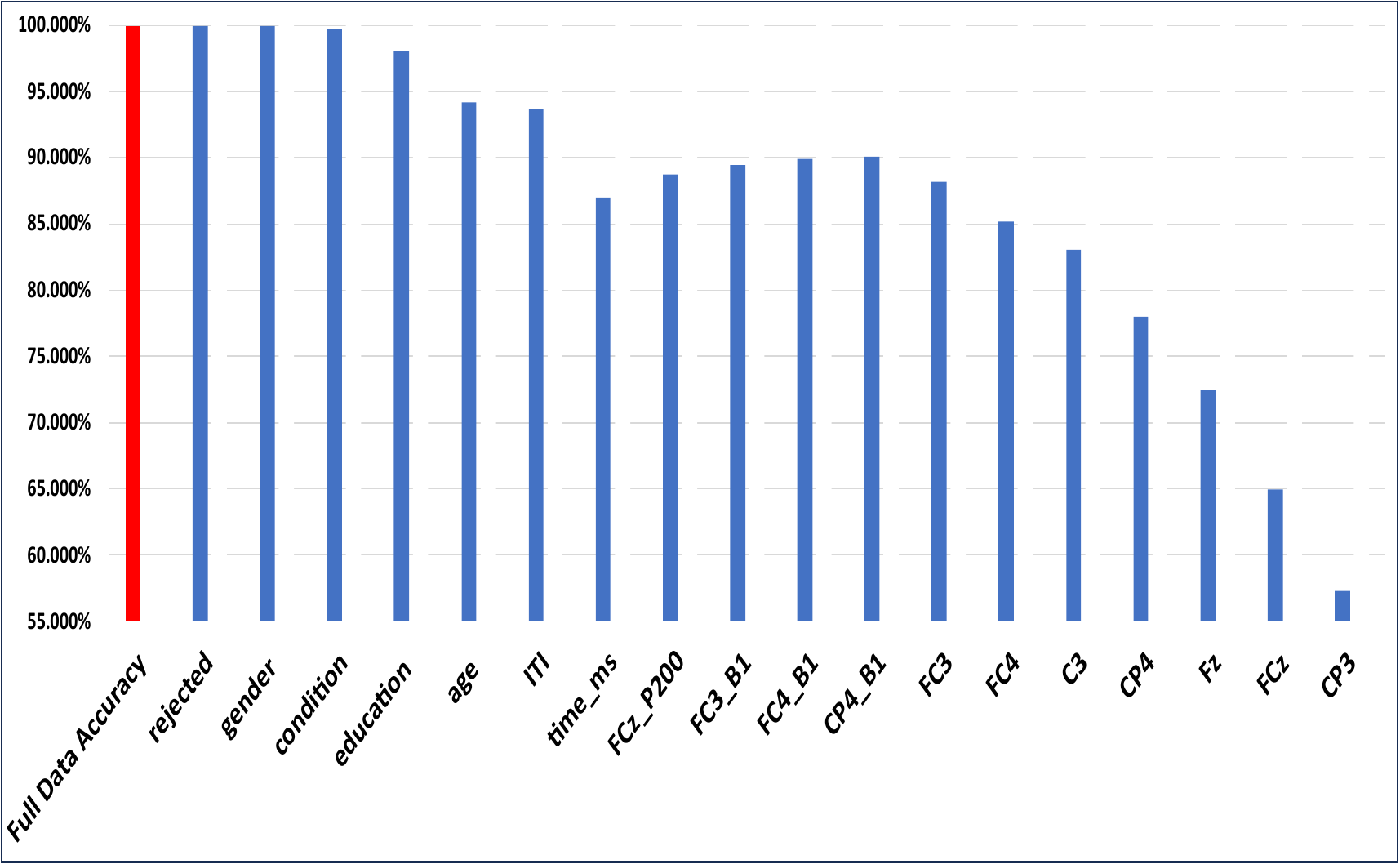}
          \caption{The accuracy resulted from incrementally removing one feature each time.}
          \label{fig:one-feature-out}
         \end{center}
        \end{figure}

Figure \ref{fig:one-feature-out} shows the accuracy changes when features were removed one by one, starting from the feature with the highest entropy until only one feature remained (incremental approach). The demographic data was the first to be removed due to its high entropy. Notably, when the age feature was removed, the accuracy dropped more significantly compared to other demographic variables.

Interestingly, the accuracy slightly increased after removing a set of 21 features, which included: ITI, $time_ms$, $FC3_B0$, $C3_B0$, $CP3_B0$, $C4_B0$, $CP4_B0$, $FC4_B0$, $Fz_B0$, $FCz_B0$, $Cz_B0$, $CP3_N100$, $Fz_N100$, $FC4_N100$, $C4_N100$, $CP4_N100$, $FC3_N100$, $FCz_N100$, $C3_N100$, $Cz_N100$, $Cz_P200$, $C4_P200$, $Fz_P200$, $FC3_P200$, $CP4_P200$, $C3_P200$, $FCz_P200$, $FC4_P200$, $CP3_P200$, $CP3_B1$, $FC3_B1$, $Fz_B1$, $Cz_B1$, $C4_B1$, $FCz_B1$, $C3_B1$, $FC4_B1$, and $CP4_B1$.

This increase in accuracy was relative to the precision of the feature removed just before rather than the original precision. Of the 51 total features, removing the majority (30 features) resulted in a noticeable decrease in accuracy. When the final set of features—C4, FC3, FC4, C3, CP4, Fz, FCz, and CP3—were removed, accuracy sharply declined, highlighting their importance. These features, deleted last, had the lowest entropy levels.

\item\textbf{Incremental one column remove experiment}


\begin{figure}[t]
        \footnotesize
         \begin{center} 
          \includegraphics[scale=0.25]{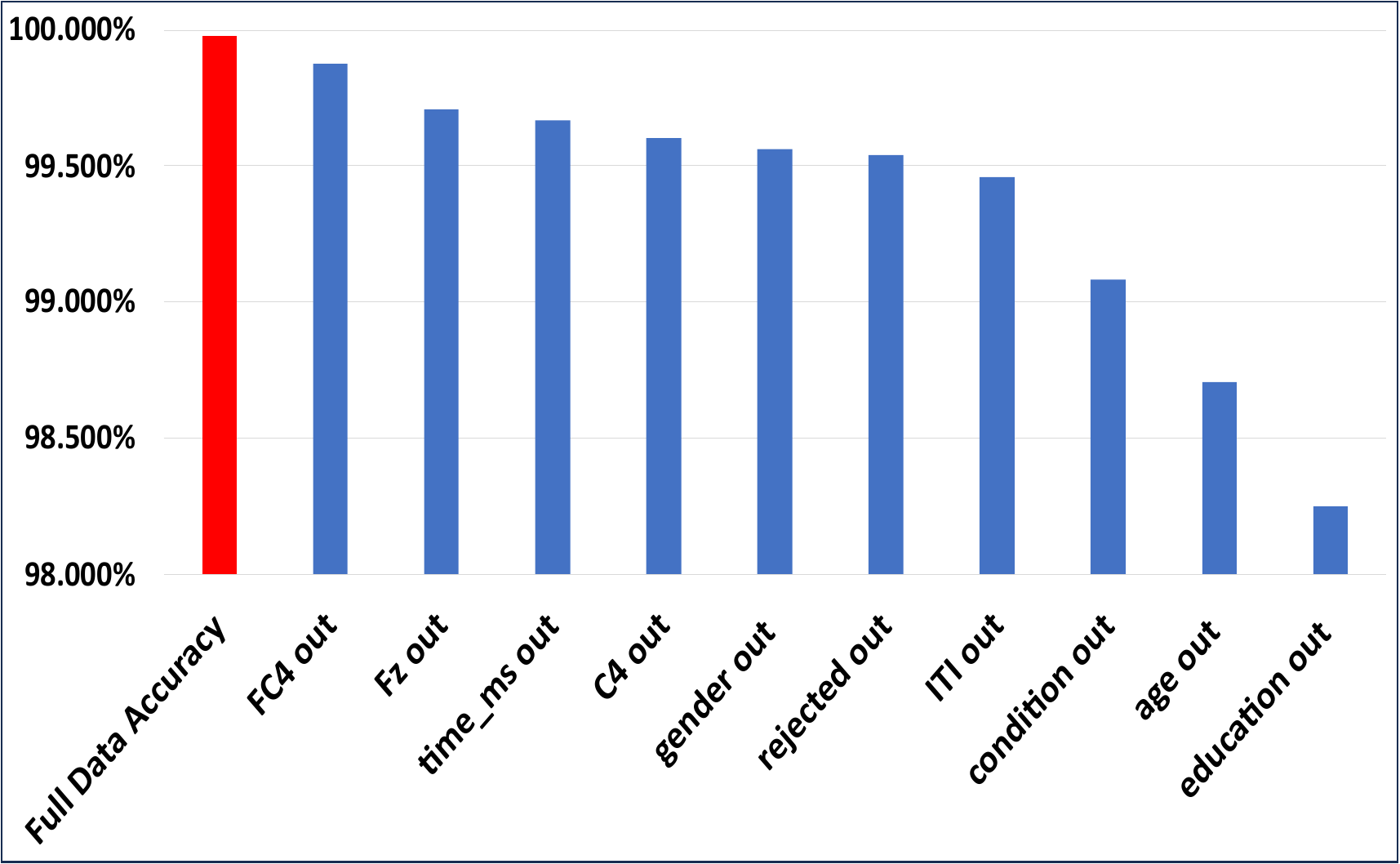}
          \caption{The accuracy results from removing one column from the original dataset each time.}
          \label{fig:Incremental_remove}
         \end{center}
        \end{figure}

Figure \ref{fig:Incremental_remove} illustrates the precise outcome of removing one feature from the entire collection of features on each occasion. The best accuracy achieved was 99.875\% after removing the FC4 feature. However, overall, it did not enhance the initial level of accuracy. On the other hand, eliminating the education feature gave us the lowest accuracy in this experiment, 98.249\%.  This indicates that education data is essential in diagnosing SZ disorders. 

Both methodologies' findings indicate that accuracy did not improve when features were removed from the data. Nevertheless, deleting one feature at a time yielded superior outcomes than gradually removing them based on their entropy ranking.
\end{itemize}

\section{Discussion}\label{sec:discussion}

This study presents a machine learning model designed to accurately classify individuals diagnosed with schizophrenia (SZ) and a control group without any mental health conditions. The model utilises electroencephalogram (EEG) data along with key demographic variables such as age, gender, and education level. Additionally, event-related potential (ERP) data is incorporated into the model. Key features, including the amplitude and latency of ERP components (N100, P200, P300), were extracted from specific time windows (e.g., 100–300 ms post-stimulus). These features were organised into a matrix, where each row represented a trial and each column contained data from various electrodes. This matrix was then used to train machine learning models aimed at classifying cognitive states. The model’s performance was evaluated using accuracy, and results indicated that certain ERP components played a significant role in predicting SZ.

Furthermore, this study highlights the importance of integrating ERP data into the training dataset, showing that the inclusion of ERP features significantly improved classification accuracy. When combined with EEG data and demographic variables, ERP data enhanced the model’s ability to reflect the cognitive symptoms associated with schizophrenia more accurately.

Their accuracy underscores the reliability of Machine Learning models, demonstrating their ability to enhance the quality of medical treatment substantially. The complexity of diseases and related data necessitates the dependability and prospective confidence of machine learning (ML) models in generating significant insights from current data.

The approach of ML tools is complex and challenging to understand. Their superior performance, on the other hand, makes them intriguing tools for automated issue prediction. Researchers need help deciding on the best ML techniques: interpretability and simplicity. Although some models are pretty predictive, they take time to grasp. Therefore, researchers have to reach a balance between accuracy and simplicity when interpreting the model.

Integration of various data will increase prediction accuracy, as we noticed when adding ERP data to the merged trial data. ML models were recently applied to utilise big data for knowledge extraction. This is an important avenue for future studies on successfully predicting SZ. 


 
In both experement that were performed on the data (incremental one feauter remove and One feature out experiments) the removal of feauters containing signals results in a reduction in the accuracy of a machine learning model, as these feauters provide critical information necessary for precise predictions. Without these key features, the model loses important patterns that are essential for distinguishing between different classes or conditions, thereby impairing its predictive performance and leading to diminished accuracy. 

A significant decline in accuracy was observed upon removing the C4, FC3, FC4, C3, CP4, Fz, FCz, and CP3 features from the dataset, underscoring their critical importance. Each electrode is located in a specific location on the head to record data from a specific location of the brain, and it is known that each part of the brain has its own function, so these electrodes will be associated with the function of this area. For example, electrodes such as C3 and C4 are fundamental in studying motor functions and movement-related potentials \cite{Sun2017}, FC3 and FC4 are essential for cognitive processing, working memory, and attention. CP3 and CP4 are pivotal in understanding sensory integration and proprioception \cite{Sun2017}. At the same time, Fz is associated with emotional regulation \cite{Marzbani2023}, inhibitory control, and decision-making while FCz play a role in studying error monitoring, executive functions, and cognitive control \cite{Cavanagh2009}.

These factors' significance for identifying SZ indicates deficiencies in those areas. Their removal disrupts the model's ability to make accurate predictions, validating their neurophysiological significance and emphasising their necessity in studies focusing on motor control, cognition, sensory processing, and emotional regulation of SZ patients.

\section{Conclusions}
\label{sec:Conclusions}

{This research highlights the efficacy of machine learning (ML) as a reliable method for predicting schizophrenia (SZ) disease. Our study has proved the efficiency of machine learning in uncovering electroencephalogram (EEG) patterns that contribute to symptoms of SZ. The researchers built a classifier based on machine learning techniques that used elements extracted from EEG data, such as event-related potential (ERP) and demographic variables. The results emphasise the efficacy of machine learning in the prediction of SZ conditions via the identification of different EEG patterns associated with SZ. The developed machine learning model, which integrated ERP, demographic characteristics, and EEG data, demonstrated proficiency in distinguishing individuals diagnosed with SZ from those deemed healthy controls. Machine learning approaches have shown encouraging results in the early diagnosis of people with SZ, giving significant therapeutic welfare. This technique allows for timely intervention during the early stages of sickness, significantly impacting clinical results. Investigating large-scale data approaches has enormous potential for future research into SZ. This necessitates exploiting massive amounts of data stored in vast databases and employing machine learning algorithms. Incorporating these approaches can significantly increase our understanding and competence in diagnosing and treating SZ.}

\section{Acknowledgments}
The researcher would like to thank the King Fahd University of Petroleum and Minerals for supporting the experiment through the high school research program (Hxplore) and Brain Roach for providing his EEG dataset. The National Institute of Mental Health funded the first data gathering (NIMH project number R01MH058262).

\bibliographystyle{acl}


\inputencoding{latin2}
\bibliography{library}
\inputencoding{utf8}

\end{document}